\documentclass[conference,a4paper]{APSIPA2025}
\usepackage{amsmath, todonotes}
\usepackage{graphicx}
\usepackage{multirow}
\usepackage{threeparttable}
\usepackage{algorithmic,textcomp,amsfonts}
\usepackage{url,gensymb,xcolor,booktabs,soul}
\usepackage{color,todonotes}
\usepackage[backend=biber,style=ieee,]{biblatex}
\usepackage{geometry}
\usepackage{fancyhdr}

\fancypagestyle{firststyle}{
  \fancyhf{}
  \fancyhead[C]{2025 Asia Pacific Signal and Information Processing Association Annual Summit and Conference (APSIPA ASC)}
  \fancyfoot[C]{\thepage}
}

\begin{document}

\title{Efficient Generative Adversarial Networks for Color Document Image Enhancement and Binarization Using Multi-scale Feature Extraction}

\author{
\authorblockN{
Rui-Yang~Ju\authorrefmark{1} and
KokSheik~Wong\authorrefmark{2} and
Jen-Shiun~Chiang\authorrefmark{3} 
}

\authorblockA{
\authorrefmark{1}
Graduate Institute of Networking and Multimedia, National Taiwan University, Taipei City, Taiwan}

\authorblockA{
\authorrefmark{2}
School of Information Technolog, Monash University Malaysia, Bandar Sunway, Malaysia}

\authorblockA{
\authorrefmark{3}
Department of Electrical and Computer Engineering, Tamkang University, New Taipei City, Taiwan}

E-mail: jryjry1094791442@gmail.com; wong.koksheik@monash.edu; jsken.chiang@gmail.com
}

\maketitle
\thispagestyle{firststyle}
\pagestyle{fancy}

\begin{abstract}
The outcome of text recognition for degraded color documents is often unsatisfactory due to interference from various contaminants. 
To extract information more efficiently for text recognition, document image enhancement and binarization are often employed as preliminary steps in document analysis. 
Training independent generative adversarial networks (GANs) for each color channel can generate images where shadows and noise are effectively removed, which subsequently allows for efficient text information extraction.
However, employing multiple GANs for different color channels requires long training and inference times.
To reduce both the training and inference times of these preliminary steps, we propose an efficient method based on multi-scale feature extraction, which incorporates Haar wavelet transformation and normalization to process document images before submitting them to GANs for training.
Experiment results show that our proposed method significantly reduces both the training and inference times while maintaining comparable performances when benchmarked against the state-of-the-art methods.
In the best case scenario, a reduction of 10\% and 26\% are observed for training and inference times, respectively, while maintaining the model performance at 73.79 of Average-Score metric.
The implementation of this work is available at \url{https://github.com/RuiyangJu/Efficient_Document_Image_Binarization}.
\end{abstract}

\section{Introduction}
Document image enhancement and binarization play important roles in document analysis, significantly impacting subsequent stages of the recognition process and layout analysis. 
For instance, color-degraded documents often suffer from various types of contaminants, such as paper yellowing, text fading, and page bleeding~\cite{sun2016blind,kligler2018document}.
These degradations seriously affect the accuracy of technologies such as Optical Character Recognition (OCR) and document image understanding.

\begin{figure}[t]
\centering
\small
\includegraphics[width=0.9\columnwidth]{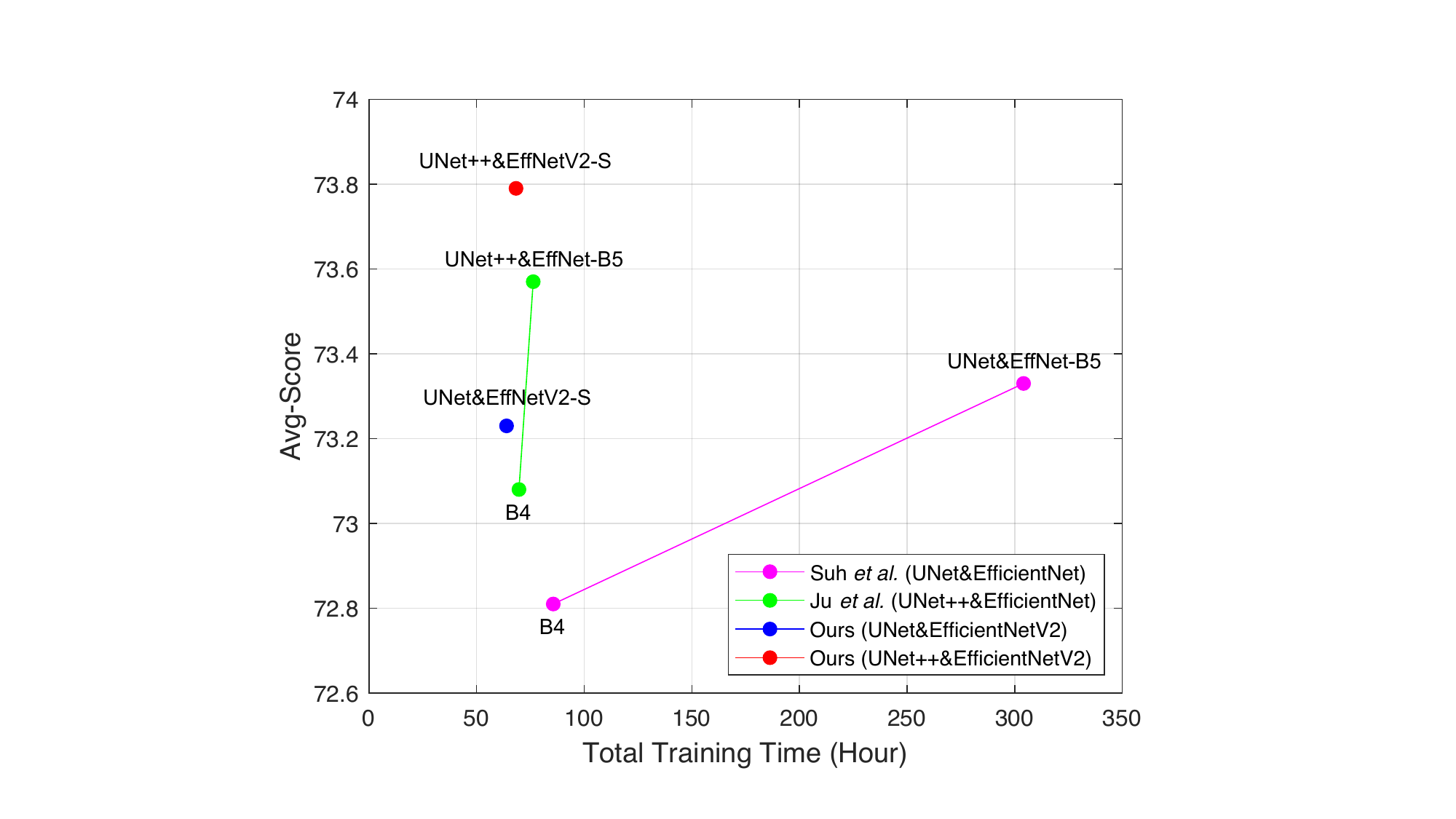}
\vskip 0.1in
\setlength{\tabcolsep}{3pt}{
\begin{tabular}{llccc}
\toprule
Method & Model & ASM$\uparrow$ & Train$\downarrow$ & Infer$\downarrow$ \\ \midrule
Suh \emph{et al.} \cite{suh2022two} & UNet\&EffNet-B5 & 73.33 & 304.12h & 0.82h \\
Ju \emph{et al.} \cite{ju2024three} & UNet++\&EffNet-B5 & 73.57 & 76.29h & 1.04h \\
Ours & UNet\&EffNetV2-S & 73.23 & 63.91h & \textbf{0.68h} \\ 
Ours & UNet++\&EffNetV2-S & \textbf{73.79} & \textbf{68.43h} & 0.77h \\ \bottomrule
\end{tabular}}
\caption{Graph of Avg-Score metric vs. Total Training Time (measured on (H)-DIBCO datasets using single NVIDIA GeForce RTX 4090 GPU).}
\label{fig:intro}
\end{figure}

Although existing state-of-the-art (SOTA) GAN-based methods~\cite{suh2022two,ju2024three} have achieved excellent performance on benchmark datasets, they do not consider training and inference times, despite these two metrics are critical in practical applications.
Our experiments reveal that these methods suffer from long training and inference times.
To address this issue, we present an efficient method including novel generators, discriminators, and loss functions for document image binarization that significantly reduces both the training and inference times while maintaining the model performance, and the results are summarized in Fig.~\ref{fig:intro}. 
Our contributions are as follows:
(a)~We are the first to introduce training and inference times as evaluation metrics, which have not been considered in existing SOTA GAN-based methods;
(b)~We adopt the Average-Score metric (ASM) to provide a more comprehensive assessment because we discover cases where PSNR value cannot correctly reflect a model's performance, and; 
(c)~Our proposed method outperforms SOTA methods in terms of model performance (as measured by ASM), training, and inference times by designing novel generators, discriminators, and loss functions.

\begin{figure*}[t]
\centering
\includegraphics[width=\linewidth]{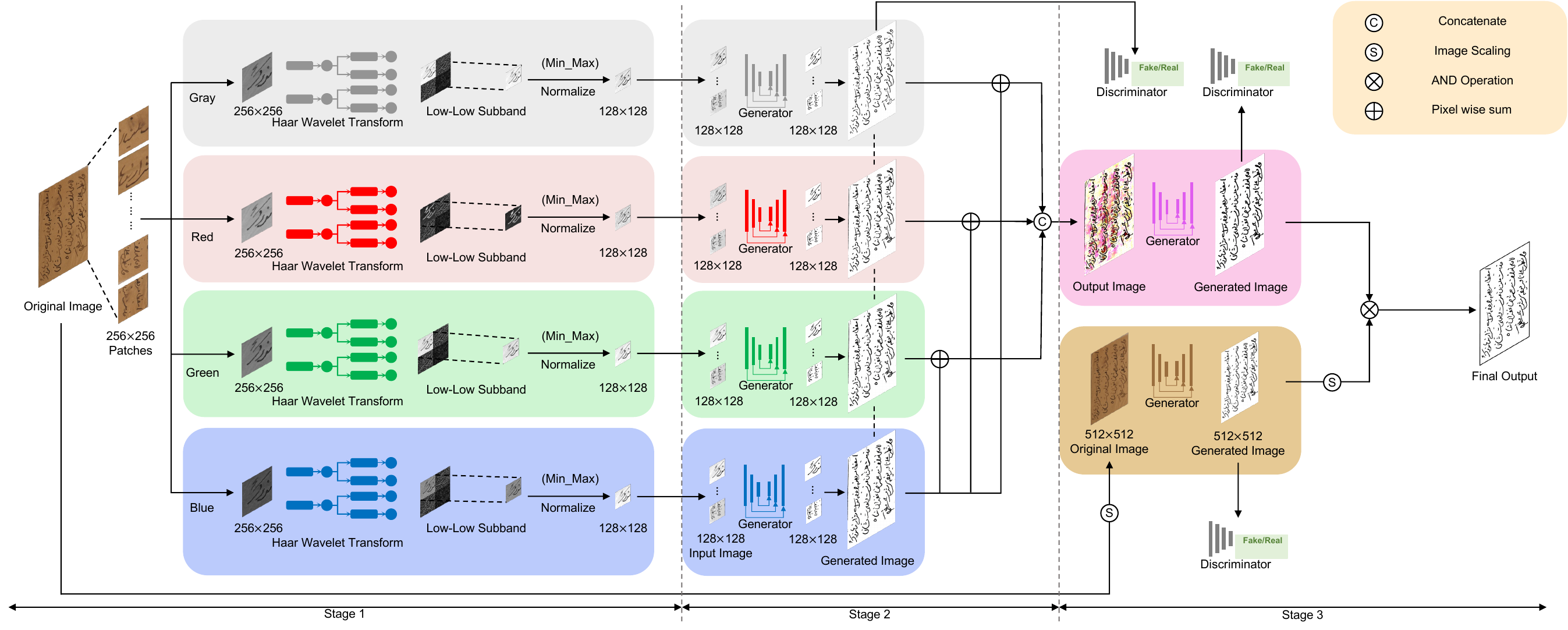}
\caption{The novel three-stage network architecture of the proposed method. 
Stage~1: document image processing, Stage~2: document image enhancement, and Stage~3: document image binarization.}
\label{fig:overall}
\end{figure*}

\section{Related Work}
Document image binarization has advanced with the introduction of fully convolutional networks (FCNs). 
Tensmeyer~\emph{et al.}~\cite{tensmeyer2017document} formulated binarization as a pixel classification learning task and utilized FCNs for this task. 
Inspired by UNet~\cite{ronneberger2015u}, Peng~\emph{et al.}~\cite{peng2017using} proposed a convolutional encoder-decoder model to perform binarization. 
He~\emph{et al.}~\cite{he2019deepotsu} proposed DeepOtsu, which initially utilized convolutional neural networks (CNNs) for document image enhancement and subsequently applied Otsu's method for document image binarization. 
In addition, Zaragoza et al.~\cite{calvo2019selectional} employed a selective autoencoder method to parse document images and subsequently binarizing them using global thresholding.

The introduction of GANs~\cite{goodfellow2020generative} has enabled the generation of binarized document images. 
Two SOTA methods based on GANs have recently been developed for document image enhancement and binarization tasks. 
Specifically, Suh~\emph{et al.}~\cite{suh2022two} proposed a novel two-stage GAN method using six improved CycleGANs~\cite{zhu2017unpaired} for color document image binarization. 
In Suh \emph{et al.}'s method, the generator consists of UNet~\cite{ronneberger2015u} with EfficientNet~\cite{tan2019efficientnet}, while the discriminator employs Pix2Pix GAN~\cite{isola2017image}. 
Ju~\emph{et al.}~\cite{ju2024three} introduced a novel three-stage GAN method based on the two-stage network architecture and employed six improved CycleGANs~\cite{zhu2017unpaired}, with an enhanced generator using UNet++~\cite{zhou2019unet++}. 
Although these methods consistently outperform SOTA models on DIBCO datasets, both of them overlook training and inference times, which is important in practical applications.

\section{Proposed Method}
\subsection{Network Architecture}
Fig.~\ref{fig:overall} presents the proposed efficient GAN method, which has a three-stage network architecture.
In Stage~1, the original color document image is divided into non-overlapping patches. 
Each patch is then split into four single-channel images (i.e., red, green, blue, and gray), because training models on different color channels tend to generate better results. 
To reduce the training time, we apply HWT and normalization to resize the image patch size from 256 $\times$ 256 to 128 $\times$ 128.

In Stage~2, we design four generators with the encoder-decoder architecture, using UNet++~\cite{zhou2019unet++} with EfficientNetV2-S~\cite{tan2021efficientnetv2} as the backbone.
Each single-channel image is fed into an independent generator for individual training. 
To standardize the generated outputs, all independent generators share the same discriminator.
Specifically, we use the improved PatchGAN~\cite{zhu2017unpaired} as the discriminator, applying instance normalization to all layers except the first layer, because including instance normalization in the first layer would normalize the color information, which is not what we aimed for.

In Stage~3, multi-scale GANs are utilized for both local and global binarization to enhance the distinction between text and background. 
The input of Stage~3 (i.e., the output of Stage~2) is an image of the same size as the original input image and is fed into an independent generator that produces the output images of local binarization ($B_l$).
In addition, the original input image is scaled to 512 $\times$ 512 pixels using Nearest Neighbour Interpolation and fed into an independent generator, and the output images of global binarization ($B_g)$ are generated. 
The final output $B$ is the pixel-wise summation of the local and global binarization results ($B =  B_l + B_g$).

\subsection{Loss Function}
Since the convergence of the loss function is unstable during the GANs training process~\cite{goodfellow2020generative}, to stabilize the loss function convergence of GANs in the proposed method, we apply Wasserstein Generative Adversarial Network with Gradient Penalty (WGAN-GP)~\cite{gulrajani2017improved} to the objective function for the model training. 
Since the goal of document image binarization is to classify each pixel into two categories (namely, text and background), we use binary cross-entropy (BCE) loss instead of $L1$ loss employed in the original method~\cite{isola2017image}. 
In addition, Galdran~\emph{et al.}~\mbox{\cite{galdran2022optimal}} demonstrated that combining BCE and Dice loss functions enhances segmentation performance at both the pixel and regional levels. 
Since better segmentation performance at the regional level leads to generated text of greater completeness, 
we use the improved WGAN-GP objective loss function, which includes both BCE loss and Soft Dice loss~\mbox{\cite{milletari2016v}} expressed below:
\begin{equation}
\begin{aligned}
\mathbb{L}_G = &-\mathbb{E}_{x}[D(G(x),x)] + \lambda_1 \mathbb{L}_{BCE}(G(x),y) \\
&+ \lambda_2 \mathop{\mathbb{L}_{Soft Dice}}(G(x),y);
\label{eq:theta_G}
\end{aligned}
\end{equation}

\begin{equation}
\begin{aligned}
\mathbb{L}_D = &-\mathbb{E}_{x,y}[D(y,x)] + \mathbb{E}_{x}[D(G(x), x)] \\
&+ \alpha \mathbb{E}_{x, \hat{y}\sim P_{\hat{y}}}[(\Arrowvert \nabla_{\hat{y}} D(\hat{y}, x) \Arrowvert_2 - 1 )^2].
\label{eq:theta_D}
\end{aligned}
\end{equation}

Here, $x$ is the input image, $G(x)$ is the generated image, and $y$ is the ground-truth image. 
$\lambda_1$ and $\lambda_2$ control the relative importance of different loss terms, while $\alpha$ denotes the gradient penalty coefficient. 
The discriminator $D$ is trained for minimizing $\mathbb{L}_D$ to distinguish between ground-truth and generated images, while the generator $G$ aims to minimize $\mathbb{L}_G$.

\section{Experiments}
\subsection{Datasets and Evaluation Metrics}
\label{sec:datasets}
To ensure a fair comparison between the proposed method and the SOTA methods~\cite{suh2022two,ju2024three}, we adopt the same strategy as in~\cite{suh2022two,ju2024three} to construct the training set. 
The training set comprises images from DIBCO 2009, H-DIBCO 2010, H-DIBCO 2012, Bickley Diary (BD), Persian Heritage Image Binarisation Dataset (PHIBD), and Synchromedia Multispectral Ancient Document Images (SMADI)~\cite{hedjam2013historical,gatos2009icdar,pratikakis2010h,deng2010binarizationshop,ayatollahi2013persian}.
The testing set consists of images from DIBCO 2011, DIBCO 2013, H-DIBCO 2014, H-DIBCO 2016, DIBCO 2017, H-DIBCO 2018, and DIBCO 2019~\cite{ntirogiannis2014icfhr2014,pratikakis2018icdar2018,pratikakis2019icdar2019}.

For quantitative comparison, four classical metrics are employed, namely, f-measure (FM), pseudo-f-measure (p-FM), peak signal-to-noise ratio (PSNR), and distance reciprocal distortion (DRD). 
When comparing the performance of different methods, there are cases where our model's FM and p-FM values reach the SOTA level, but our PSNR value is lower than that of other methods. 
Inspired by Jemni~\mbox{\emph{et al.}~\cite{jemni2022enhance}}, we adopt the Average-Score metric (ASM) to evaluate the overall performance of each method more comprehensively:
\begin{equation}
ASM := \dfrac{FM\!+\!p\!\!-\!\!FM\!+\!PSNR\!+\!(100\!-\!DRD)}{4}.
\end{equation}
Note that in ASM, precision and recall have a greater impact on the value than PSNR, which we consider reasonable.
This is because, for methods utilizing GANs to generate binarized images, the focus should be on the overall correctness of the generated image rather than on individual pixels. 

\subsection{Implementation Details}
To ensure a fair comparison of performances, we utilize the same dataset and data augmentation techniques for our proposed method and the SOTA methods~\cite{suh2022two,ju2024three}. 
In Stage~1, the original input images are split into 256 $\times$ 256 patches, to match the size of the images from the ImageNet~\cite{deng2009imagenet} dataset, considering that we will use the pre-trained model based on this dataset.
Data augmentation is then employed to expand the training samples, with sampling scales set at 0.75, 1, 1.25, 1.5, and rotation by 270$\degree$, resulting in a total of 120,174 training image patches. 
For global binarization (Stage~3), the input images are directly resized to 512 $\times$ 512 and subjected to horizontal and vertical flipping, as well as rotation by 90$\degree$, 180$\degree$, and 270$\degree$, resulting in 804 training images.

To avoid the influence of hardware differences on model performance, all methods are trained using a single NVIDIA RTX4090 GPU. 
In addition, all methods are implemented in Python using PyTorch as the framework.
The backbone networks in all methods use weights pre-trained on the ImageNet~\cite{deng2009imagenet} dataset to enhance efficiency in model training.
The training parameter settings are largely similar for Stage~2 and Stage~3, except for the number of training epochs, i.e., 10 epochs for Stage~2 and 150 epochs for Stage~3.
We choose Adam optimizer to train models and set the initial learning rate to 2$\times10^{-4}$. 
In addition, we configure generators with $\beta_1 =$ 0.5 and discriminators with $\beta_2 =$ 0.999. 

\begin{table}[t]
\centering
\small
\caption{Training and inference times taken by the SOTA method (Baseline) and after apply HWT and normalization (Ours).}
\setlength{\tabcolsep}{4pt}{
\begin{tabular}{lcccccc}
\toprule
Method & \begin{tabular}[c]{@{}c@{}}Stage2\\ Train\end{tabular} & \begin{tabular}[c]{@{}c@{}}Stage2\\ Predict\end{tabular} & \begin{tabular}[c]{@{}c@{}}Stage3\\ Top\end{tabular} & \begin{tabular}[c]{@{}c@{}}Stage3\\ Bottom\end{tabular} & \begin{tabular}[c]{@{}c@{}}Total\\ Train\end{tabular} & \begin{tabular}[c]{@{}c@{}}Total\\ Infer\end{tabular} \\ \midrule
Baseline & 332.28h & 3.56h & 47.47h & 1.63h & 384.95h & 1.12h \\
Ours & 11.60h & 3.45h & 47.47h & 1.39h & 63.91h & 0.68h \\ \midrule
Baseline & 465.28h & 3.94h & 52.88h & 1.76h & 523.86h & 1.19h \\
Ours & 14.12h & 3.63h & 49.29h & 1.39h & 68.43h & 0.77h \\ \bottomrule
\multicolumn{7}{l}{\footnotesize{Above two methods use Model~A, and below two use Model~B.}} \\
\end{tabular}}
\label{tab:ablation_time}
\end{table}

\begin{table}[t]
\centering
\small
\caption{PSNR (dB) of images resized using different methods: Interpolation/HWT/HWT\&Normalization (Ours).}
\setlength{\tabcolsep}{2pt}{
\begin{tabular}{lccccccc}
\toprule
Method & 2009 & 2010 & 2012 & BD & PHIBD & SMADI & Mean Values \\ \midrule
Bicubic & {\color{blue}71.45} & {\color{blue}72.22} & 71.67 & 64.29 & 69.58 & {\color{blue}69.88} & {\color{blue}69.85} \\
Bilinear & 70.94 & 72.16 & 71.46 & 64.07 & {\color{blue}69.71} & 69.86 & 69.70 \\
Area & 70.94 & 72.16 & 71.46 & 64.07 & {\color{blue}69.71} & 69.86 & 69.70 \\
Nearest & 70.95 & 72.04 & 71.59 & 64.20 & 69.69 & 69.83 & 69.72 \\
Lanczos & 71.42 & {\color{blue}72.22} & {\color{blue}71.69} & {\color{blue}64.30} & 69.58 & {\color{red}69.89} & {\color{blue}69.85} \\
HWT & 62.65 & 67.11 & 59.67 & 53.76 & 58.00 & 59.48 & 60.11 \\
\textbf{Ours} & {\color{red}71.77} & {\color{red}72.74} & {\color{red}72.85} & {\color{red}64.44} & {\color{red}70.76} & 69.44 & {\color{red}70.34} \\ \bottomrule
\multicolumn{8}{l}{\footnotesize{The best and 1st runner up performances are in {\color{red}red} and {\color{blue}blue}, respectively.}}
\end{tabular}}
\label{tab:psnr}
\end{table}

\begin{table*}[t]
\centering
\small
\caption{\small Quantitative comparison (ASM: FM/p-FM/PSNR/DRD, Total Training Time, Total Inference Time) of the proposed method and SOTA methods for document image enhancement and binarization on DIBCO datasets.}
\setlength{\tabcolsep}{6.2pt}{
\begin{tabular}{ccccccccc}
\toprule
Method & Model & FM$\uparrow$ & p-FM$\uparrow$ & PSNR$\uparrow$ & DRD$\downarrow$ & ASM$\uparrow$ & Total Train$\downarrow$ & Total Inference$\downarrow$ \\ \cmidrule(r){1-2} \cmidrule(r){3-6} \cmidrule(r){7-9}
Suh \emph{et al.}~\cite{suh2022two} & UNet\&EfficientNet-B4 & 87.95 & 89.01 & 19.10dB & 4.83 & 72.81 & 85.61h & {\color{blue}0.74h} \\
Suh \emph{et al.}~\cite{suh2022two} & UNet\&EfficientNet-B5 & 88.56 & 89.90 & {\color{red}19.31dB} & {\color{red}4.46} & 73.33 & 304.12h & 0.82h \\ \cmidrule(r){1-2} \cmidrule(r){3-6} \cmidrule(r){7-9}
Ju \emph{et al.}~\cite{ju2024three} & UNet++\&EfficientNet-B4 & 88.14 & 89.71 & 19.09dB & 4.64 & 73.08 & 69.68h & 0.91h \\
Ju \emph{et al.}~\cite{ju2024three} & UNet++\&EfficientNet-B5 & {\color{blue}89.13} & {\color{blue}90.35} & {\color{blue}19.30dB} & 4.49 & {\color{blue}73.57} & 76.29h & 1.04h \\ \cmidrule(r){1-2} \cmidrule(r){3-6} \cmidrule(r){7-9}
\textbf{Ours} & UNet\&EfficientNetV2-S & 88.83 & 89.87 & 19.07dB & 4.86 & 73.23 & {\color{red}63.91h} & {\color{red}0.68h} \\
\textbf{Ours} & UNet++\&EfficientNetV2-S & {\color{red}89.69} & {\color{red}90.78} & 19.15dB & {\color{blue}4.45} & {\color{red}73.79} & {\color{blue}68.43h} & 0.77h \\ \bottomrule
\multicolumn{9}{l}{\footnotesize{The best and 1st runner up performances are in {\color{red}red} and {\color{blue}blue}, respectively.}}
\end{tabular}}
\label{tab:comparison_performance}
\end{table*}

\begin{figure*}[t]
\centering
\small
\includegraphics[width=2\columnwidth]{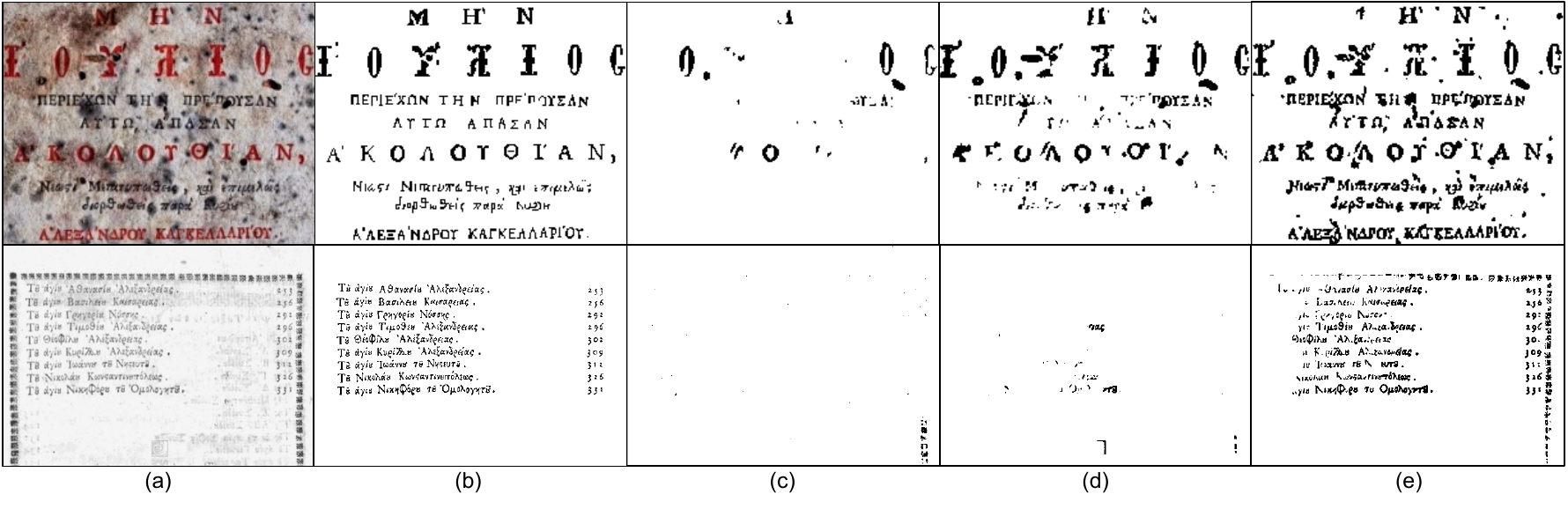}
\setlength{\tabcolsep}{13pt}{
\begin{tabular}{cccccccc}
\toprule
\multirow{2}{*}{Method} & \multirow{2}{*}{Model} & \multicolumn{3}{c}{The first row images:} & \multicolumn{3}{c}{The second row images:} \\ 
 &  & FM$\uparrow$ & p-FM$\uparrow$ & PSNR$\uparrow$ & FM$\uparrow$ & p-FM$\uparrow$ & PSNR$\uparrow$ \\ \cmidrule(r){1-2} \cmidrule(r){3-5} \cmidrule(r){6-8}
Blank Image & -- & -- & -- & 10.90dB & -- & -- & 14.19dB \\
Suh \emph{et al.}~\cite{suh2022two} & UNet\&EfficientNet-B4 & 26.09 & 19.90 & 11.39dB & 0.60 & 0.60 & 14.00dB \\
Ju \emph{et al.}~\cite{ju2024three} & UNet\&EfficientNet-B4 & 60.39 & 56.35 & 12.19dB & 10.05 & 9.32 & 14.23dB \\
Ours & UNet\&EfficientNetV2-S & 69.44 & 69.88 & 11.75dB & 56.99 & 56.51 & 14.08dB \\ \bottomrule
\end{tabular}}
\caption{\small Representative visualized results from the test set: (a) input image, (b) ground-truth, (c) Suh~\emph{et al.}~\cite{suh2022two}, (d) Ju~\emph{et al.}~\cite{ju2024three}, and (e) ours.}
\label{fig:result}
\vspace{-1em}
\end{figure*}

\begin{table*}[t]
\centering
\small
\caption{\small Ablation study on each improvement step. The checkmark ($checkmark$) indicates that a specific configuration is in use.
The last row records the results for the proposed method.}
\setlength{\tabcolsep}{6.2pt}{
\begin{tabular}{cccccccc}
\toprule
\begin{tabular}[c]{@{}c@{}}Generator:\\EfficientNetV2-S\end{tabular} & \begin{tabular}[c]{@{}c@{}}Generator:\\UNet++\end{tabular} & \begin{tabular}[c]{@{}c@{}}Discriminator:\\InstanceNorm\end{tabular} & \begin{tabular}[c]{@{}c@{}}Generator Loss Function:\\$D(G(z))$+$\lambda_1$BCE+$\lambda_2$DICE\end{tabular} & \begin{tabular}[c]{@{}c@{}}Image Processing:\\HWT\&Norm\end{tabular} & ASM & \begin{tabular}[c]{@{}c@{}}Total\\Train\end{tabular} & \begin{tabular}[c]{@{}c@{}}Total\\Inference\end{tabular} \\ \cmidrule(r){1-5} \cmidrule(r){6-8}
 & \checkmark & \checkmark & \checkmark & \checkmark & 73.79 & 112.74h & 1.21h \\
\checkmark &  & \checkmark & \checkmark & \checkmark & 73.23 & 63.91h & 0.68h \\
\checkmark & \checkmark &  & \checkmark & \checkmark & 73.45 & 61.24h & 0.89h \\
\checkmark & \checkmark & \checkmark &  & \checkmark & 73.58 & 70.52h & 0.91h \\
\checkmark & \checkmark & \checkmark & \checkmark &  & 73.81 & 523.86h & 1.19h \\ 
\checkmark & \checkmark & \checkmark & \checkmark & \checkmark & 73.79 & 68.43h & 0.77h \\ \bottomrule
\end{tabular}}
\label{tab:ablation}
\end{table*}

\subsection{Multi-scale Feature Extraction}
To reduce total training and inference times, our work proposes to resize both input and the corresponding ground-truth images for GANs training by half. 
To illustrate the effectiveness of HWT and normalization in Stage~1, we consider two GAN models, namely: 
Model~A: UNet~\cite{ronneberger2015u} with EfficientNetV2-S~\cite{tan2021efficientnetv2}, and Model~B: UNet++~\cite{zhou2019unet++} with EfficientNetV2-S~\cite{tan2021efficientnetv2}. 
Table~\ref{tab:ablation_time} records the time taken for each stage, as well as the total training and inference times of different methods.
Two configurations are compared: ``with the application of HWT and normalization in Stage~1'' and ``without (i.e., the \emph{baseline} where split patches are directly supplied to the GANs)''. 
Here, the total training time is the sum of the time taken for each stage, and the total inference time is the total time taken to generate images for all test sets. 
It can be seen that, for both models, the total training time is reduced when HWT and normalization are applied. 
Specifically, when HWT and normalization are applied, the training time is reduced from 384.95h to 63.91h for Model~A, and from 523.86h to 68.43h for Model~B.
Similarly, the total inference time is reduced from 1.12h to 0.68h for Model~A, and from 1.19h to 0.77h for Model~B. 
This demonstrates that the use of HWT and normalization can significantly reduce the total training and inference times.

We also explore other image-resizing techniques, including interpolation-based algorithms such as bicubic, bilinear, area, nearest neighbor, and Lanczos. 
We implement these techniques using the open-source computer vision library (OpenCV) to downscale all input images and the corresponding ground truth images from 256 $\times$ 256 to 128 $\times$ 128.
Furthermore, we employ the ``HWT'', and ``HWT and normalization (HWT\&Norm)'' methods. 
It is noteworthy that the resized images from all these methods are not binarized, which cannot be used to calculate the PSNR values directly with the corresponding ground-truth (binary) images. 
Therefore, we first apply global binarization to these resized images, and then compute the PSNR values.
We evaluate the impact of different image resizing techniques on six training sets by calculating the PSNR values (against the corresponding ground-truth images), and we compute the mean PSNR values for all images. 
The results are recorded in Table~\ref{tab:psnr}. 
We observe that the mean PSNR value achieved by ``HWT'' method is 60.11dB, indicating that images reduced directly using HWT have low similarity with the corresponding ground-truth images. 
In addition, the mean PSNR values for resized images produced by different interpolation methods are all below 70dB. 
However, the mean PSNR value for images processed by ``HWT and normalization'' reaches 70.34dB, which confirms that the images obtained by this method are closer to the corresponding ground-truth images at the pixel level. 
In conclusion, the results demonstrate that our ``HWT and normalization'' method is more effective than other interpolation-based image-resizing techniques for document image enhancement and binarization.

\subsection{Comparison with SOTA Methods}
We compare our proposed methods with the SOTA methods using GANs~\cite{suh2022two,ju2024three} for document image enhancement and binarization, where the results are shown in Table~\ref{tab:comparison_performance}.
Considering that the total training time for methods using UNet~\cite{ronneberger2015u} or UNet++~\cite{zhou2019unet++} with EfficientNet-B5~\cite{tan2019efficientnet} is already longer than that of our proposed method, we do not further compare the methods using EfficientNet-B6, as it is against the goal of reducing training and inference times.

We can see that our proposed method using UNet++~\cite{zhou2019unet++} with EfficientNetV2-S~\cite{tan2021efficientnetv2} achieves the highest ASM of 73.79. 
It requires a total training time of 68.43h, which is also thesecond shortest time.
It is faster than UNet++\&EfficientNet-B5 (76.29h) that yields the second highest ASM.
Furthermore, the total inference time of our method is 0.77h, which is notably lower than 1.04h as required by Ju~\emph{et al.}'s method~\cite{ju2024three} using UNet++ with EfficientNet-B5, representing a reduction of approximately 26\%.
Moreover, the proposed method using UNet with EfficientNetV2-S obtains the shortest total training time and inference time of 63.91h and 0.68h, respectively.
Although our achieved ASM value of 73.23 is not the highest, when compared to Suh~\emph{et al.}'s method~\cite{suh2022two} (using UNet with EfficientNet-B5) that yields 73.33 ASM, the training time is reduced from 304.12h to 63.91h, which is a remarkable decrease of approximately 78\%.
Overall, the experiment results demonstrate the efficiency and competitive performance of our proposed method.

Next, we compare the results achieved by all benchmark methods for each evaluation metric. 
Our method achieves the highest FM and p-FM values of 89.69 and 90.79, respectively, while maintaining lower total training and inference times than the method with the second highest FM and p-FM values. 
For the DRD metric, our method achieves the second highest value, but with a significantly reduced total training time of 68.43h compared to 304.12h taken by the method with the highest DRD value. 
Although our method does not achieve the highest PSNR, this metric does not directly reflect the model performance in document image enhancement and binarization.
To validate this statement, we randomly select two images from the test set for visual inspection.
As shown in Fig.~\ref{fig:result}, our method generates more complete foreground information. 
However, due to the high contamination of the document image, some noise is inevitable while generating more content. 
In contrast, Suh~\emph{et al.}'s method~\cite{suh2022two} and Ju~\emph{et al.}'s method~\cite{ju2024three} generate less content. 
It is such because 
the background is white, and PSNR favors methods generating less content, i.e., they will have higher PSNR values. 
To put things into context, a blank image (i.e., all pixels set to white) yields a PSNR of 14.19dB, which is higher than that of our proposed method (14.08dB).
However, it is obvious that the binarized image generated by our method is closer to the ground-truth image than the blank image. 
These observations support our claim that a higher PSNR value is non-indicative of better model performance, and our method can successfully generate more textual information.

\subsection{Ablation Study}
To evaluate the contribution of each enhancement in our proposed method, we gradually replace or remove each improvement step and observe the impact on performance.
Table~\ref{tab:ablation} summarizes the results under various configurations.
To ensure experimental fairness, all experiments were conducted using the same dataset and hyperparameter settings.

Specifically, replacing UNet~\cite{ronneberger2015u} with UNet++~\cite{zhou2019unet++} in the generator and adding instance normalization to the discriminator improve model performance, with a slight increase in training time.
Replacing EfficientNet-B5~\cite{tan2019efficientnet} with EfficientNetV2-S~\cite{tan2021efficientnetv2} in the generator reduces both training and inference times, while the new loss function further improves performances.
Finally, employing HWT and normalization for multi-scale feature extraction significantly reduces training time from 523.86h to 68.43h, representing an 87\% decrease, with a drop of only 0.02 in ASM.
Overall, each improvement objectively contributes to either performance gains or reductions in training and inference times.

\section{Conclusion and Future Work}
Degraded color document image enhancement and binarization are important steps in document analysis. 
The current SOTA methods based on GANs can generate satisfactory document binarization results, but suffer from long 
training and inference times.
To address this drawback, we propose a three-stage GAN method using HWT and normalization for multi-scale feature extraction, which greatly reduces the total training and inference times. 
Furthermore, novel generators, discriminators, and a loss function are designed to further improve the performance of our proposed method. 
Experiment results on benchmark datasets demonstrate that the proposed method not only achieves superior model performance but also significantly reduces the total training and inference times in comparison to SOTA methods.

As future exploration, we can combine document image binarization and document image understanding for practical applications, especially for ancient documents or historical artifacts. 
The applications could include real-time translation, summarization, and retrieval of related documents/materials.

\section*{Acknowledgment}
This work is supported by National Science and Technology Council of Taiwan, under Grant Number: NSTC 112-2221-E-032-037-MY2.

\printbibliography
\end{document}